\documentclass{article}
\pdfoutput=1

\usepackage{arxiv}

\usepackage[utf8]{inputenc} 
\usepackage[T1]{fontenc}    
\usepackage{hyperref}       
\usepackage{url}            
\usepackage{booktabs}       
\usepackage{amsfonts}       
\usepackage{nicefrac}       
\usepackage{microtype}      
\usepackage{graphicx}
\usepackage{siunitx}
\usepackage[numbib]{tocbibind}
\usepackage{makecell}
\usepackage{float}

\title{RAGulator: Lightweight Out-of-Context Detectors for Grounded Text Generation}

\author{
 Ian Poey \\
  OCBC \\
  Singapore \\
  \texttt{ianpoey@ocbc.com} \\
  \And
 Jiajun Liu \\
  OCBC \\
  Singapore \\
  \texttt{jiajunliu@ocbc.com} \\
  \And
 Qishuai Zhong \\
  OCBC \\
  Singapore \\
  \texttt{qishuaizhong@ocbc.com} \\
  \And
 Adrien Chenailler \\
  OCBC \\
  Singapore \\
  \texttt{adrienchenailler@ocbc.com} \\
}

\begin{document}
\maketitle
\begin{abstract}
Real-time detection of out-of-context LLM outputs is crucial for enterprises looking to safely adopt RAG applications. In this work, we train lightweight models to discriminate LLM-generated text that is semantically out-of-context from retrieved text documents. We preprocess a combination of summarisation and semantic textual similarity datasets to construct training data using minimal resources. We find that DeBERTa is not only the best-performing model under this pipeline, but it is also fast and does not require additional text preprocessing or feature engineering. While emerging work demonstrates that generative LLMs can also be fine-tuned and used in complex data pipelines to achieve state-of-the-art performance, we note that speed and resource limits are important considerations for on-premise deployment.
\end{abstract}

\section{Introduction}
In enterprise settings, Generative AI has received widespread adoption as a tool to uplift employees’ productivity \cite{brachman2024how}. Enterprises require a high degree of factuality of generated answers in popular language tasks, such as question-answering (QA) and summarisation. 

Two methods can enable Large Language Models (LLMs) to interact with users based on industry-specific knowledge. The first method involves fine-tuning through strategies such as SFT, RLHF \cite{ouyang2022training} and preference optimisation \cite{rafailov2024dpo, lu2024discopop}. Although able to improve performance on domain knowledge, knowledge gained from fine-tuning can quickly become outdated \cite{gekhman2024does}, therefore making maintenance costly. The alternative is Retrieval Augmented Generation (RAG) \cite{lewis2021retrieval}, which is better suited for tasks requiring evolving knowledge, such as integration of the latest industry news. However, both methods do not fully circumvent the inherent limitation of LLMs – hallucination, which manifests in inconsistent or fabricated claims \cite{huang2023survey} that can be subtle and phrased confidently even if factually incorrect \cite{li2024drowzee}. For highly sensitive working environments such as financial institutions, the inability to ensure faithful LLM outputs can be one of the biggest limitations to widespread adoption of LLM applications \cite{thealanturinginstitute2024impact}.

To address this barrier to enterprise adoption, we narrow the scope of “hallucination” and focus only on hallucinations that render the LLM response semantically inconsistent with the provided context. This is commonly known as \textit{faithfulness hallucination} \cite{huang2023survey, es2023ragas, saadfalcon2024ares}, but is sometimes also referred to by terms such as \textit{contextual hallucination} \cite{chuang2024lookback}, or a lack of \textit{grounded factuality} \cite{bespoke2024llmaggrefactblog} or \textit{support} \cite{belyi2024luna}; for simplicity and to clearly underscore the nature of such hallucination, we adopt the term \textit{out-of-context} (OOC). We consider any LLM-generated response to a RAG prompt as semantically OOC if any part of the response is ungrounded based on the retrieved context alone, even if it is otherwise factual according to world knowledge. In contrast, an \textit{in-context} response is one where every claim embedded in the response can be inferred solely from the retrieved context.

Approaches to mitigate hallucination can be broadly classified into 2 types – black-box and grey-box methods. \textbf{Black-box methods} employ strong generative LLMs to assess if a candidate answer is hallucinated, and these LLMs are usually augmented with various prompting or fine-tuning techniques. Examples of black-box methods include RARR \cite{gao2023rarr}, WikiChat \cite{semnani2023wikichat}, FreshPrompt \cite{vu2023fresh}, SelfCheckGPT \cite{manakul2023selfcheckgpt}, RAGAS \cite{es2023ragas}, ChainPoll \cite{friel2023chainpoll}, and Lynx \cite{ravi2024lynx}. However, black-box approaches that rely on strong closed-source LLM judges are less suitable for enterprises that are constrained by budget and/or data privacy requirements. \textbf{Grey-box methods} are alternatives which aim to detect hallucination through a proxy metric or model. Grey-box approaches either assess final/hidden LLM states, such as FLARE \cite{jiang2023flare} and Lookback Lens \cite{chuang2024lookback}, or use a score computed by an independent discriminative model of lower complexity, such as SummaC \cite{laban2021summac}, AlignScore \cite{zha2023alignscore}, HHEM \cite{vectara2024hhem}, and Luna \cite{belyi2024luna}.

In this work, we propose RAGulator, a series of lightweight OOC detectors for RAG applications. We use a simple data generation pipeline to create a training dataset which simulates both OOC and in-context RAG prompts. This is gathered from public datasets originally constructed for various NLP tasks. Furthermore, we compare 2 types of grey-box "non-native" discriminative models – fine-tuned BERT-based classifiers and ensemble meta-classifiers trained on numerical features derived from text. Generative labelling with an LLM annotator is employed where necessary to adapt the training dataset for fine-tuning of the BERT classifiers. We show that while a large LLM can show good agreement with human annotation in labelling data for BERT classifier fine-tuning, our predictive models outperform the same LLM on the OOC detection task by up to 19\% on AUROC and 17\% on F1 score (deberta-v3-large), highlighting the need for specialised models for OOC detection.

\begin{figure}[H]
    \centering
    \includegraphics[width=1\linewidth]{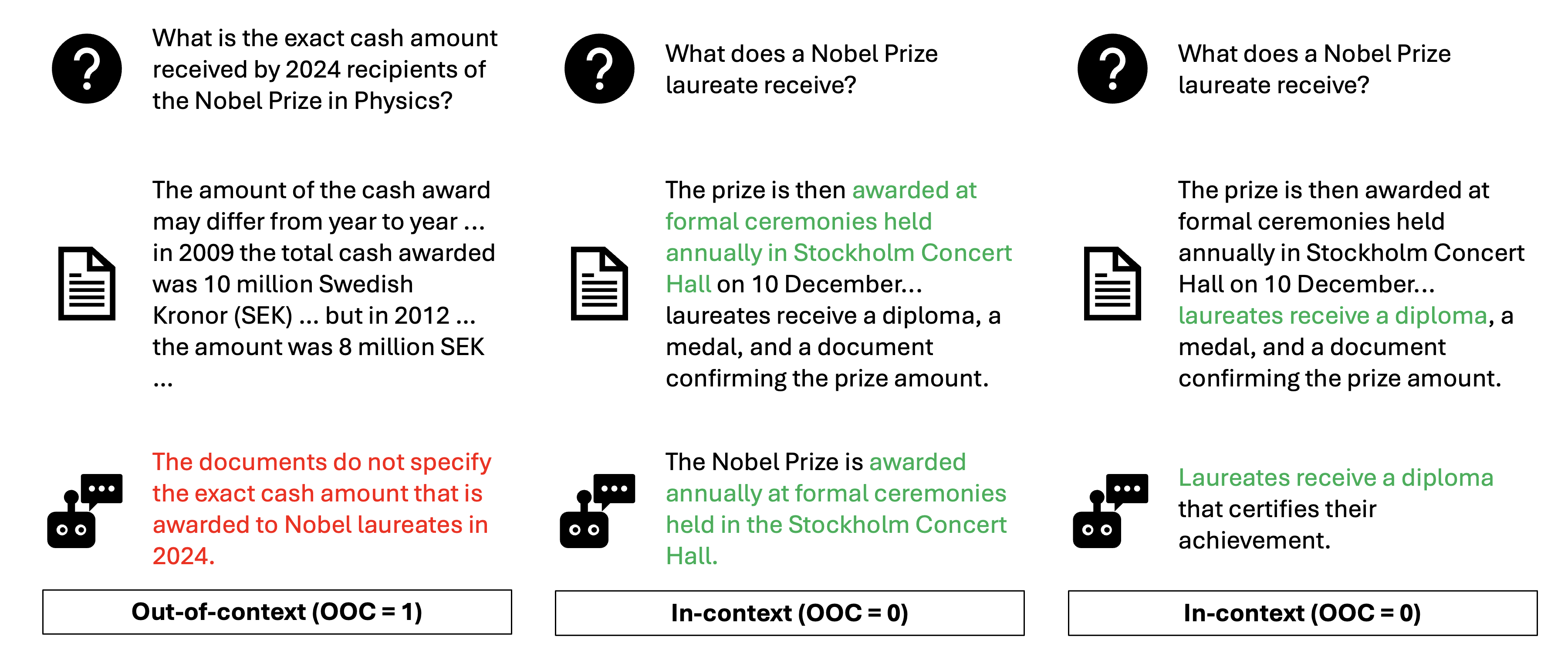}
    \caption{Example LLM responses from a RAG system}
    \label{fig:ooc-example}
\end{figure}

\section{Method}

\subsection{Problem formulation}
We train a lightweight grey-box discriminator to detect semantically OOC LLM-generated sentences from a RAG system. Figure \ref{fig:ooc-example} illustrates examples of OOC and in-context sentences. In the OOC example (left column), the LLM is unable to find the answer within the retrieved documents, and the generated statement is not grounded in context despite directly answering the question. On the other hand, the in-context statements are clearly grounded in the documents. For in-context sentences, their individual relevance to the question can be low (middle column) or high (right column).

We formulate the problem of semantic OOC as such. In our formulation we closely follow the setup by Tang et al. \cite{tang2024minicheck}, but exclude the generalisation made to accommodate post-hoc grounding scenarios. Any RAG prompt $Prompt(\mathcal{D}, x)$ to an LLM contains an associated set of retrieved documents $\mathcal{D} = \{D_1,...,D_{|\mathcal{D}|}\}$ and a question $x$, to which the LLM responds with text that can be broken into an unordered set of sentences $\mathbf{c} = \{c_1,...,c_{|\mathbf{c}|}\}$. We treat each sentence $c_i$ as an independent candidate claim to be verified. To detect OOC candidates, we define a discriminator
\begin{center}
  $M(\mathcal{D}, c_i) \in \{0,1\}$
\end{center}
that classifies each candidate as \verb|out-of-context|, 1, or \verb|in-context|, 0, according to the set of retrieved documents $\mathcal{D}$.

In our formulation, we ignore the relevance of the candidate claim $c_i$ to the question $x$. A claim can be generated such that it reiterates the question statement as an assertive statement (hence being relevant to the question) but it is ungrounded with respect to the retrieved documents; such cases should be considered OOC.

Certain model architectures (e.g. BERT) may face limitations on the amount of text that can be accepted by the model in one pass. This would require the set of documents $\mathcal{D}$ to be arbitrarily re-split into $J$ text chunks $\{D_1,...,D_{J}\}$ of varying sizes not more than the model-defined limit. In this case, the overall discriminator can be defined as
\begin{center}
  $\min_jM(\mathcal{D}_j, c_i) \in \{0,1\}$
\end{center}
implying that if and only if the claim $c_i$ is OOC with respect to every text chunk $D_j$, then it can be considered OOC overall.

\subsection{Dataset curation}
We construct a dataset by adapting publicly available datasets, sampling and preprocessing them to simulate LLM-generated sentences and RAG-retrieved contexts of various lengths. This curation is straightforward – it only involves sentence tokenisation (using SpaCy) and random sampling, without any application of machine learning. Datasets belonging to summarisation (extractive and abstractive) and semantic textual similarity tasks were selected for this purpose.
\paragraph{Summarisation task datasets.}
In its original form, each row in the dataset is an abstract-article pair. Preprocessing was done by randomly pairing abstracts with unrelated articles to create "hallucinated" OOC pairs, then sentencizing the abstracts to create one example for each abstract sentence. The following datasets were selected:
\begin{itemize}
    \item BBC \cite{greene2006icml}: this extractive text summarisation dataset contains 2,225 news articles from 5 topical areas published by BBC between 2004 and 2005, with their respective summaries.
    \item CNN/Daily Mail \cite{nallapati2016abstractive}: this extractive and abstractive text summarisation dataset (version 3.0.0) contains over 300,000 unique English-language news articles written by journalists at CNN and the Daily Mail, with their respective summaries.
    \item PubMed \cite{cohan-etal-2018-discourse}: this dataset contains summaries for about 133,000 documents pulled from the scientific repository PubMed.com, where the average document and summary lengths are much longer than both BBC and CNN/Daily Mail.
\end{itemize}
\paragraph{Semantic textual similarity (STS) datasets.}
In its original form, each row in the dataset is a sentence pair, accompanied by a label indicating if the pair of sentences are similar. Preprocessing was done by inserting random sentences from the datasets to one of the sentences in the pair to simulate a long "context". The original labels were mapped to our definitions to indicate if the pair is an OOC pair (dissimilar sentence pair). The following datasets were selected:
\begin{itemize}
    \item Microsoft Research Paraphrase Corpus (MRPC) \cite{dolan-brockett-2005-automatically}: the MRPC dataset contains 5,801 pairs of sentences extracted from English-language newswire articles available online, along with human-annotated binary labels indicating whether each pair captures a semantic equivalence relationship. The semantically-equivalent indicator was mapped to \verb|in-context| if present and \verb|out-of-context| if absent.
    \item Stanford Natural Language Inference (SNLI) \cite{snli:emnlp2015}: The SNLI corpus (version 1.0) comprises 570,000 human-written English sentence pairs manually labelled for balanced three-way classification. Labelling was performed by up to five human annotators, and original labels are in the set \{\verb|entailment|, \verb|contradiction|, \verb|neutral|\}. We disregarded sentence pairs labelled as \verb|neutral| and those where not all annotators were in agreement, while mapping the labels \verb|entailment| and \verb|contradiction| to \verb|in-context| and \verb|out-of-context| respectively.
\end{itemize}
In our eventual curated dataset, we ensure each simulated sentence has a length of between 5-100 tokens, and that each simulated context has a length of between 100-5000 tokens. The OOC class ratio is maintained throughout training and holdout evaluation. We sample only from the original training and test splits provided by the authors to form training and held-out evaluation data respectively. We only include PubMed, MRPC and SNLI datasets in our evaluation split as they are generally more challenging datasets.

\subsection{Generative labelling}
Since BERT models have a standard 512-token limit, in order to fine-tune such models for the OOC detection task, our fine-tuning data should consist of sentence-context pairs with a combined tokenised length of 512 tokens maximum. We manipulate our dataset into this format via generative labelling.

\begin{figure}
    \centering
    \includegraphics[width=1\linewidth]{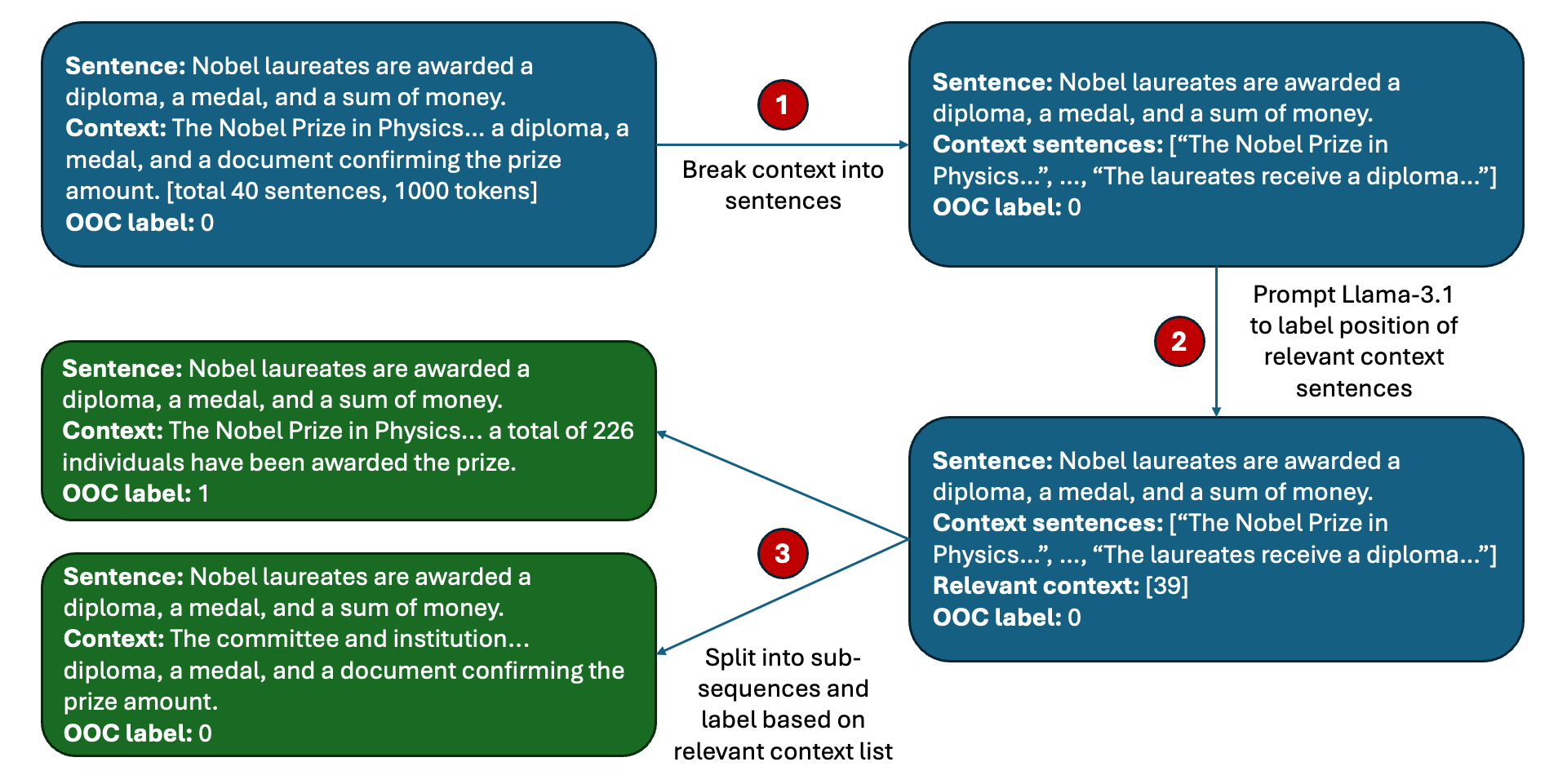}
    \caption{Illustrated flow of the generative labelling process}
    \label{fig:gen-label}
\end{figure}

For sentence-context examples derived from summarisation datasets and labelled as not hallucinated ($\approx$34\% of dataset), the LLM sentence is relevant to at least one part of the context, but the exact position is unknown. As shown in Figure \ref{fig:gen-label}, we utilise Llama-3.1-70b-Instruct to label each sentence by prompting it to return the positions of context sentences that are relevant. With this output, we split the sentence-context example into windows of tokenised sub-sequences (i.e. [CLS] + sentence tokens + [SEP] + context tokens) that are each no longer than 512 tokens, and label the sub-sequences based on whether the sub-sequence contains a relevant context sentence.

\subsection{Feature engineering}
We utilise classic machine translation metrics to compare response (candidate) and context (reference) alongside semantic relation between two texts approximated using distance metrics calculated from encoder models. From preliminary investigations, we found the distribution of each feature to be different between in-context LLM sentences and OOC LLM sentences.

\begin{itemize}
    \item \textbf{Precision score:} defined as the fraction of words in the response that also appeared in the context \cite{melamed-etal-2003-precision}. We apply the precision score after text preprocessing and de-duplication of words in both candidate and reference.
    \item \textbf{Unigram perplexity:} we construct a probability distribution (i.e. dictionary of token frequencies) from the context's unigram tokens after text preprocessing, then calculate sum negative log likelihood and normalise by the number of candidate unigrams.
    \item \textbf{Bigram perplexity:} same as unigram perplexity but calculated for bigrams.
    \item \textbf{Maximum embedding similarity score:} we use bge-small-en-v1.5 by the Beijing Academy of Artificial Intelligence (BAAI) \cite{xiao2024bge} to compute maximum pairwise similarity between embeddings of the response and each context sentence. The embedding model was initialised with pretrained weights.
    \item \textbf{Maximum reranker relevance score:} we use bge-reranker-base, also by BAAI, to compute the maximum relevance score of the response against each context sentence. The reranker model was initialised with pretrained weights.
\end{itemize}

\paragraph{Text preprocessing.}
Retrieved context can contain noise due to reasons such as sub-optimal parsing of reference documents, presence of formatting characters and inconsistent grammatical or lexical structure. Therefore, we applied text preprocessing to both response and context before generating classical features. Preprocessing steps include conversion to lowercase, removal of most punctuation symbols, word stemming and stopword removal. For semantic features, we used the raw response-context pair that BERT-based models expect as input.

\paragraph{Score aggregation.}
Similarity and relevance scores were aggregated by maximum – as long as the candidate is relevant to one context sentence, the candidate should be considered grounded in context, therefore capturing the extent to which the candidate is supported.

\subsection{Models}
We apply the described feature engineering technique on our prepared dataset to train two meta-classifiers of LightGBM and Random Forest architectures respectively. We also apply the described generative labelling technique on our prepared dataset to train two BERT-based classifiers: deberta-v3-large initialised with default weights, and xlm-roberta-large initialised with weights from bge-reranker-large (i.e. fine-tuning the reranker). 

Training details for all model types can be found in Section \ref{appendix-hyperparams}.

\section{Evaluation}

\subsection{Data}
We evaluate RAGulator models on a curated evaluation set based on two sources:
\begin{itemize}
    \item In-distribution holdout split of our simulated RAG dataset as described in this work.
    \item Out-of-distribution collection of 184 RAG responses that were gathered and hand-annotated credit policy documents (henceforth “CP”) used by bank officers. 
\end{itemize}

\subsection{Model inference}
When running inference with BERT classifiers, we predict on each of the tokenised sub-sequences of the original sentence-context pair. We then aggregate model predictions to obtain a single prediction at the sentence level. We take the minimum probability across all sub-sequences as the overall OOC probability – as long as one of the generated sub-sequences is in-context, we consider the entire generated sentence as in-context.

We use Llama-3.1-70b-Instruct hosted on 4x H100 GPUs, with all other models hosted on 1x V100 GPU.

\section{Results}

\subsection{Generative labelling vs. human annotation}
We verify the effectiveness of generative labelling by experimenting with several prompting methods on a sample of 58 sentence-context pair examples (20 BBC, 20 CNN/Daily Mail, 18 PubMed), amounting to a total of 2,696 context sentences. Table \ref{tab:gen-label} shows the inter-rater agreement between human annotation and each Llama-3.1 annotator prompted with different methods, such as few-shot prompting, chain-of-thought prompting (COT), or both. Details on prompting can be found in Section \ref{appendix-llama-label}.

We find that the use of five-shot and COT increases the reliability of generative labelling with respect to human annotation, evidenced by the increase in Cohen's kappa ($\kappa$). Nevertheless, direct zero-shot prompting is already in substantial agreement with human observation. We eventually adopted this method for generative labelling as this method is the fastest, utilises the fewest tokens, and least likely to deviate in terms of desired LLM output structure.

\subsection{Model performance on evaluation datasets}
\paragraph{Performance comparison.}
We compare our models to Llama-3.1-70b-Instruct using direct prompting as a baseline in Table \ref{tab:result-test}. Details on prompting and extracted OOC probabilities are in Section \ref{appendix-llama-ooc}. Across the full evaluation set, our models consistently outperform Llama-3.1 in AUC metrics. The observed performance gap is even larger on the CP dataset. The best-performing model, deberta-v3-large, shows a 19\% increase in AUROC and a 17\% increase in F1 score despite being significantly smaller than Llama-3.1.

\paragraph{Inference speed and size.}
The BERT-based models are at least 630\% faster than Llama-3.1, with xlm-roberta-large achieving the fastest speed at over 26 examples (sentences) per second. In contrast, the meta-classifiers are up to 33.7\% slower than Llama-3.1 at just over 2 examples per second. However, the 70B model requires 4 H100 GPUs to achieve this observed speed. While each meta-classifier and their underlying models can be loaded on a single V100 GPU, this is not feasible for Llama-3.1 at this size.

\begin{table}
    \centering
    \begin{tabular}{lll}
    \toprule
    Prompting method & Accuracy vs. human (\%) & $\kappa$ (\%) \\
    \midrule
    0-shot & 98.5 & 77.8 \\
    0-shot, COT & 98.3 & 78.6 \\
    5-shot & 98.6 & 80.5 \\
    5-shot, COT & 98.7 & 83.6 \\
    \bottomrule
    \end{tabular}
    \caption{Measure of inter-rater agreement between human annotation and each Llama-3.1 prompting method}
    \label{tab:gen-label}
\end{table}

\begin{table}
    \centering
    \begin{tabular}{lccccccc}
    \toprule
     
    & \multicolumn{3}{c}{Test - overall} 
    & \multicolumn{3}{c}{Test - CP} \\
    \cmidrule(r){2-4} \cmidrule(r){5-7}
    Model & AUROC & AUPRC & F1 & AUROC & AUPRC & F1 & Speed (examples / s)\\
    \midrule
    Llama-3.1-70b-Instruct & 89.3 & 91.0 & 88.7 & 75.7 & 82.8 & 68.4 & 3.03 \\
    Random Forest & 94.5 & 94.6 & 86.7 & 89.6 & 85.3 & 79.7 & 2.01 \\
    LightGBM & 95.5 & 95.4 & 87.4 & 89.7 & 81.7 & 77.9 & 2.03 \\
    deberta-v3-large & \textbf{98.2} & \textbf{98.3} & \textbf{93.5} & \textbf{90.1} & \textbf{86.5} & \textbf{80.3} & 19.08 \\
    xlm-roberta-large & 97.1 & 97.2 & 91.1 & 88.4 & 86.3 & 78.6 & 26.09 \\
    \bottomrule
    \end{tabular}
    \caption{Results on overall evaluation set and in-house CP dataset}
    \label{tab:result-test}
\end{table}

\section{Discussion}
We train RAGulator through a lightweight pipeline, where the most computationally intensive step is the calculation of embedding similarity and reranker relevance scores during feature engineering. Embeddings and (optionally) relevance scores are precomputed metrics in the context retrieval process of RAG. The same embedding and reranker models used for context retrieval can be reused in the feature engineering pipeline to decrease the total resources used in the overall RAG system.

We introduce an out-of-distribution evaluation dataset not only to prevent data leakage, but also to assess if a given OOC model is suitable for our use case. We ensure that these goals are met by collecting private enterprise data of a different domain from that of the training data, so that its distribution does not overlap with the training distribution.

The main benefit that RAGulator offers is that model training and inference both require fewer resources as opposed to black-box detection methods. However, this introduces limitations that could be avenues for future research.

In framing the semantic OOC problem, we treat all OOC sentences as a single label without distinguishing between common fine-grained hallucination sub-types such as entity, relation, invented, or subjective hallucinations \cite{mishra2024fava}. To train hallucination detection models for this purpose, a high-quality training dataset must contain candidate sentences that are similar in meaning to the retrieved context, while also containing at least one hallucination sub-type. A popular direction in recent work is to use strong LLMs (e.g. GPT-4) in data pipelines to generate high-quality candidates containing hallucination sub-types. This includes error insertion, answer perturbation, and the generation of new claims from documents, rendering human annotators unnecessary for model training \cite{ravi2024lynx, tang2024minicheck, mishra2024fava, li2023halueval}. The high-quality dataset is then used in a black-box approach to fine-tune generative LLMs. While shown to have a positive impact on detection models, such pipelines and fine-tuning would not be feasible in a closed development environment constrained by either cost, computational resources, or data privacy. Future work could explore low-resource approaches to training models for fine-grained hallucination detection using high-quality, open-source generated datasets.

Furthermore, the training set for RAGulator does not include RAG use cases outside of QA and summarisation, such as code generation or data-to-text writing from tabular data. This work could be extended to such use cases by incorporating relevant high-quality training data where available.

\section{Conclusion}
In this work, we demonstrate a low-resource pipeline for data gathering and training of RAGulator, a series of small discriminative models to detect LLM-generated texts that are semantically out-of-context. We show that the tested models outperform LLM-as-a-judge zero-shot detection with Llama-3.1-70b-Instruct, a much larger state-of-the-art decoder LLM. Of these models, deberta-v3-large is not only the best-performing model, but also over 6x faster than a 70B LLM judge and does not require additional text preprocessing or feature engineering. Although emerging work has focused on LLM-as-a-judge methods or on fine-tuning of small generative LLMs to achieve state-of-the-art performance, speed and resource limits are concerns when considering to deploy such pipelines \cite{parthasarathy2024ultimate}. For enterprises that are bound by strict data security rules and resource constraints, our work presents a favourable alternative for on-premise hallucination detection.

\bibliographystyle{unsrt}
\bibliography{references}
\clearpage

\section{Appendix}
\subsection{Model training hyperparameters}
\label{appendix-hyperparams}
Table \ref{tab:hyperparams-tree} shows the hyperparameter search space used to tune the LightGBM and Random Forest meta-classifiers, while Table \ref{tab:hyperparams-bert} shows the relevant hyperparameters for the training of BERT classifiers. During BERT fine-tuning, we observed that the validation performance no longer increases after 3 epochs.

\begin{table}
    \centering
    \begin{tabular}{lcc}
    \toprule
    Hyperparameter & LightGBM & Random Forest \\
    \midrule
    max\_depth & [2,4,-1] & [1,2,3,4,5] \\
    n\_estimators & [60,100,200] & [100,325,550,775,1000] \\
    num\_leaves & [4,10,31] & - \\
    subsample & [0.8,1.0] & - \\
    \bottomrule
    \end{tabular}
    \caption{Grid search values for hyperparameter tuning of meta-classifiers}
    \label{tab:hyperparams-tree}
\end{table}

\begin{table}
    \centering
    \begin{tabular}{lc}
    \toprule
    Hyperparameter & Value \\
    \midrule
    Batch size & 16 \\
    Epochs & 3 \\
    Learning rate (base model) & \num{5e-6} \\
    Learning rate (classification head) & \num{2e-5} \\
    Warmup steps & 10\% of total training steps \\
    \bottomrule
    \end{tabular}
    \caption{Hyperparameter values used for training of BERT classifiers}
    \label{tab:hyperparams-bert}
\end{table}

\subsection{Llama-3.1 for generative labelling}
\label{appendix-llama-label}
The prompt template for zero-shot generative labelling of training examples is as follows:

\begin{minipage}{\linewidth}
  \fbox{%
    \parbox{\linewidth}{%
\#\#\# Instruction \#\#\# \\ You are a professional proofreader with an eye for detail. Given a candidate sentence and a numbered list of context sentences, your job is to identify which context sentences support the claim in the candidate sentence. A context sentence supports the claim in the candidate sentence if it is an exact match, or close to an exact match. You also know that there must be at least one supporting context sentence. Return the supporting context sentence identified by its index only. If there are multiple supporting context sentences, delimit the indexes by comma. \\ \\ \#\#\# Candidate sentence \#\#\# \\ <This is the candidate sentence> \\ \\ \#\#\# Context sentences \#\#\# \\ 0. """<This is the first context sentence>""" \\ 1. """<This is the second context sentence>""" \\ ... \\ \\ \#\#\# Response \#\#\# \\
    }%
  }%
\end{minipage}

To elicit chain-of-thought, the following line was added to the instruction while leaving all else unchanged:

\begin{minipage}{\linewidth}
  \fbox{%
    \parbox{\linewidth}{%
        Let's think step-by-step. After presenting your thought process, list down the supporting context sentences by writing "The answer is:" followed by the list.
    }%
  }%
\end{minipage}
\clearpage

The prompt template for five-shot labelling is as follows:

\begin{minipage}{\linewidth}
  \fbox{%
    \parbox{\linewidth}{%
\#\#\# Instruction \#\#\# \\ You are a professional proofreader with an eye for detail. Given a candidate sentence and a numbered list of context sentences, your job is to identify which context sentences support the claim in the candidate sentence. A context sentence supports the claim in the candidate sentence if it is an exact match, or close to an exact match. You also know that there must be at least one supporting context sentence. Return the supporting context sentence identified by its index only. If there are multiple supporting context sentences, delimit the indexes by comma. A few examples of the task will be given, followed by the actual task. \\ \\ 
\# Example 1 \# \\
\#\#\# Candidate sentence \#\#\# \\ Washington, D.C. is the capital of the US. \\ \\ \#\#\# Context sentences \#\#\# \\ 0. """Washington, D.C., formally the District of Columbia, is the capital city and federal district of the United States.""" \\ 1. """The city is on the Potomac River, across from Virginia.""" \\ 2. """The city was founded in 1791, and the 6th Congress held the first session in the unfinished Capitol Building in 1800 after the capital moved from Philadephia.""" \\ \\ \#\#\# Response \#\#\# \\ {[0]} \\ \\
<Examples 2-5> \\ \\
\# Actual task \# \\
\#\#\# Candidate sentence \#\#\# \\ <This is the candidate sentence> \\ \\ \#\#\# Context sentences \#\#\# \\ 0. """<This is the first context sentence>""" \\ 1. """<This is the second context sentence>""" \\ ... \\ \\ \#\#\# Response \#\#\# \\
    }%
  }%
\end{minipage}
\clearpage

To elicit chain-of-thought for five-shot prompting, the following prompt template was used:

\begin{minipage}{\linewidth}
  \fbox{%
    \parbox{\linewidth}{%
\#\#\# Instruction \#\#\# \\ You are a professional proofreader with an eye for detail. Given a candidate sentence and a numbered list of context sentences, your job is to identify which context sentences support the claim in the candidate sentence. A context sentence supports the claim in the candidate sentence if it is an exact match, or close to an exact match. You also know that there must be at least one supporting context sentence. Let's think step by step. After presenting your thought process, list down the supporting context sentences by writing "The answer is:" followed by the list. Return the supporting context sentence identified by its index only. If there are multiple supporting context sentences, delimit the indexes by comma. A few examples of the task will be given, followed by the actual task. \\ \\ 
\# Example 1 \# \\
\#\#\# Candidate sentence \#\#\# \\ Washington, D.C. is the capital of the US. \\ \\ \#\#\# Context sentences \#\#\# \\ 0. """Washington, D.C., formally the District of Columbia, is the capital city and federal district of the United States.""" \\ 1. """The city is on the Potomac River, across from Virginia.""" \\ 2. """The city was founded in 1791, and the 6th Congress held the first session in the unfinished Capitol Building in 1800 after the capital moved from Philadephia.""" \\ \\ \#\#\# Response \#\#\# \\ To identify the supporting context sentences, I will analyse each context sentence to see if it matches or closely matches the claim in the candidate sentence. The candidate sentence claims that "Washington, D.C. is the capital of the US." Context sentence 0 states that "Washington, D.C., formally the District of Columbia, is the capital city and federal district of the United States." This sentence is an exact match, as it also states that Washington, D.C. is the capital of the United States. Context sentence 1 mentions the location of the city, but does not mention its status as the capital. Context sentence 2 mentions the history of the city and the capital moving from Philadelphia, but does not explicitly state that Washington, D.C. is the current capital. The answer is: {[0]} \\ \\
<Examples 2-5> \\ \\
\# Actual task \# \\
\#\#\# Candidate sentence \#\#\# \\ <This is the candidate sentence> \\ \\ \#\#\# Context sentences \#\#\# \\ 0. """<This is the first context sentence>""" \\ 1. """<This is the second context sentence>""" \\ ... \\ \\ \#\#\# Response \#\#\# \\
    }%
  }%
\end{minipage}
\clearpage

\subsection{Direct prompting of Llama-3.1 for OOC detection}
\label{appendix-llama-ooc}
To perform OOC detection with Llama-3.1, we directly prompt it to return a binary prediction – 1 for OOC and 0 otherwise. The template used for prompting is shown below:

\begin{minipage}{\linewidth}
  \fbox{%
    \parbox{\linewidth}{%
\#\#\# Instruction \#\#\# \\ Given a candidate sentence and a reference text, your job is to identify if the reference text supports the claim in the candidate sentence. The reference text supports the claim in the candidate sentence if it is an exact match. The reference text may also support the claim if the claim is grounded in, or entails, the reference text. Respond with the final prediction only: 0 if the claim is supported, 1 if the claim is not supported. \\ \\ \#\#\# Candidate sentence \#\#\# \\ <This is the candidate sentence> \\ \\ \#\#\# Reference text \#\#\# \\ <This is the reference text> \\ \\ \#\#\# Response \#\#\# \\
    }%
  }%
\end{minipage}

To obtain an OOC prediction probability $P(OOC)$, we extract the log-probabilities (logprobs) of the “0” or “1” token choices in the first output token and calculate softmax of the "0" token based on these 2 choices only:
\begin{center}
  $P(OOC) = \frac{\exp(log(P(y="1")))}{\exp(log(P(y="0")))+\exp(log(P(y="1")))}$
\end{center}
where $log(P(y="1"))$ is the logprob of predicting the "1" token and $log(P(y="0"))$ is the logprob of predicting the "0" token.

\paragraph{Error handling.}
If the logprob of either token choice cannot be found in the top 10 token choices, we estimate the logprob of the missing token choice by taking the sum of the remaining 9 logprobs. In the rare scenario that neither token can be found in the top 10 token choices, we assume an OOC prediction with a probability of 1.

\end{document}